\documentclass[letterpaper, 10 pt, conference]{ieeeconf}  

\IEEEoverridecommandlockouts                              

\usepackage{graphics} 
\usepackage{epsfig} 
\usepackage{times} 
\usepackage{amsmath} 
\usepackage{amssymb}  
\usepackage{bm}
\usepackage{color}
\usepackage{cite}

\usepackage{ulem} 
\usepackage{hyperref}
\usepackage{float}
\usepackage{booktabs}
\usepackage{multirow}
\usepackage{makecell}
\usepackage{mathrsfs}
\usepackage[utf8]{inputenc}
\usepackage[caption=false,font=footnotesize]{subfig}
\usepackage{graphicx}
\usepackage{pifont}
\usepackage{soul}
\usepackage{algorithmic,algorithm}
\usepackage{bbding}

\newcounter{RNum}
\renewcommand{\theRNum}{\arabic{RNum}}
\newcommand{\Remark}{\noindent\textit{\textbf{Remark}~\refstepcounter{RNum}\textbf{\theRNum}: }}
\newcommand{\NoOne}[1]{\textcolor{red}{#1}}
\newcommand{\NoTwo}[1]{\textcolor{green}{#1}}
\newcommand{\NoThree}[1]{\textcolor{blue}{#1}}
\soulregister\NoOne7
\soulregister\NoTwo7
\soulregister\NoThree7
\soulregister\Remark7

\title{\LARGE \bf
Neural-iLQR: A Learning-Aided Shooting Method \\for Trajectory Optimization
}

\author{Zilong Cheng$^{1,*}$, Yulin Li$^{2,*}$, Kai Chen$^{2}$, Jun Ma$^{2}$, Tong Heng Lee$^{1}$ 
 \thanks{$^{*}$ indicates equal contribution.}%
\thanks{$^{1}$ Zilong Cheng and Tong Heng Lee are with the Department of Electrical and Computer Engineering, National University of Singapore, Singapore 117583 (e-mail: zilongcheng@u.nus.edu; eleleeth@nus.edu.sg).
}%
\thanks{$^{2}$ Yulin Li, Kai Chen, and Jun Ma are with The Hong Kong University of Science and Technology, China (e-mail: yline@connect.ust.hk; kchen916@connect.hkust-gz.edu.cn; jun.ma@ust.hk). }}

\begin{document}

\maketitle
\thispagestyle{empty}
\pagestyle{empty}


\begin{abstract}
Iterative linear quadratic regulator (iLQR) has gained wide popularity in addressing trajectory optimization problems with nonlinear system models. However, as a model-based shooting method, it relies heavily on an accurate system model to update the optimal control actions and the trajectory determined with forward integration, thus becoming vulnerable to inevitable model inaccuracies. Recently, substantial research efforts in learning-based methods for optimal control problems have been progressing significantly in addressing unknown system models, particularly when the system has complex interactions with the environment. Yet a deep neural network is normally required to fit substantial scale of sampling data. In this work, we present Neural-iLQR, a learning-aided shooting method over the unconstrained control space, in which a neural network with a simple structure is used to represent the local system model. In this framework, the trajectory optimization task is achieved with simultaneous refinement of the optimal policy and the neural network iteratively, without relying on the prior knowledge of the system model. Through comprehensive evaluations on two illustrative control tasks, the proposed method is shown to outperform the conventional iLQR significantly in the presence of inaccuracies in system models.
\end{abstract}

\section{Introduction} \label{sec:intro}
The last decade has witnessed substantial achievements in the context of trajectory optimization, pervading different application domains including unmanned aerial vehicles (UAVs) \cite{minimum-snap}, autonomous driving \cite{Autonomous-Vehicle-Recurrent-Spline-Optimization}, quadrupeds \cite{Gait-optimizaition-phase-based}, mobile manipulators \cite{constraint-DDP}, etc. However, it is still an open and challenging problem on the generation of a satisfying trajectory in complex scenarios, especially when the state/control space is inherently high-dimensional, the system model is nonlinear, and the non-smooth contact and constraints are introduced in most of the robot systems \cite{constraint-DDP,Fast-planner,two-stage}. For such problems, the model-based method and the learning-based method are extensively investigated.

\begin{figure}[t]	
	\centering
        \vspace{3pt}
	\includegraphics[trim=50 40 100 100, clip,width=0.8\linewidth]{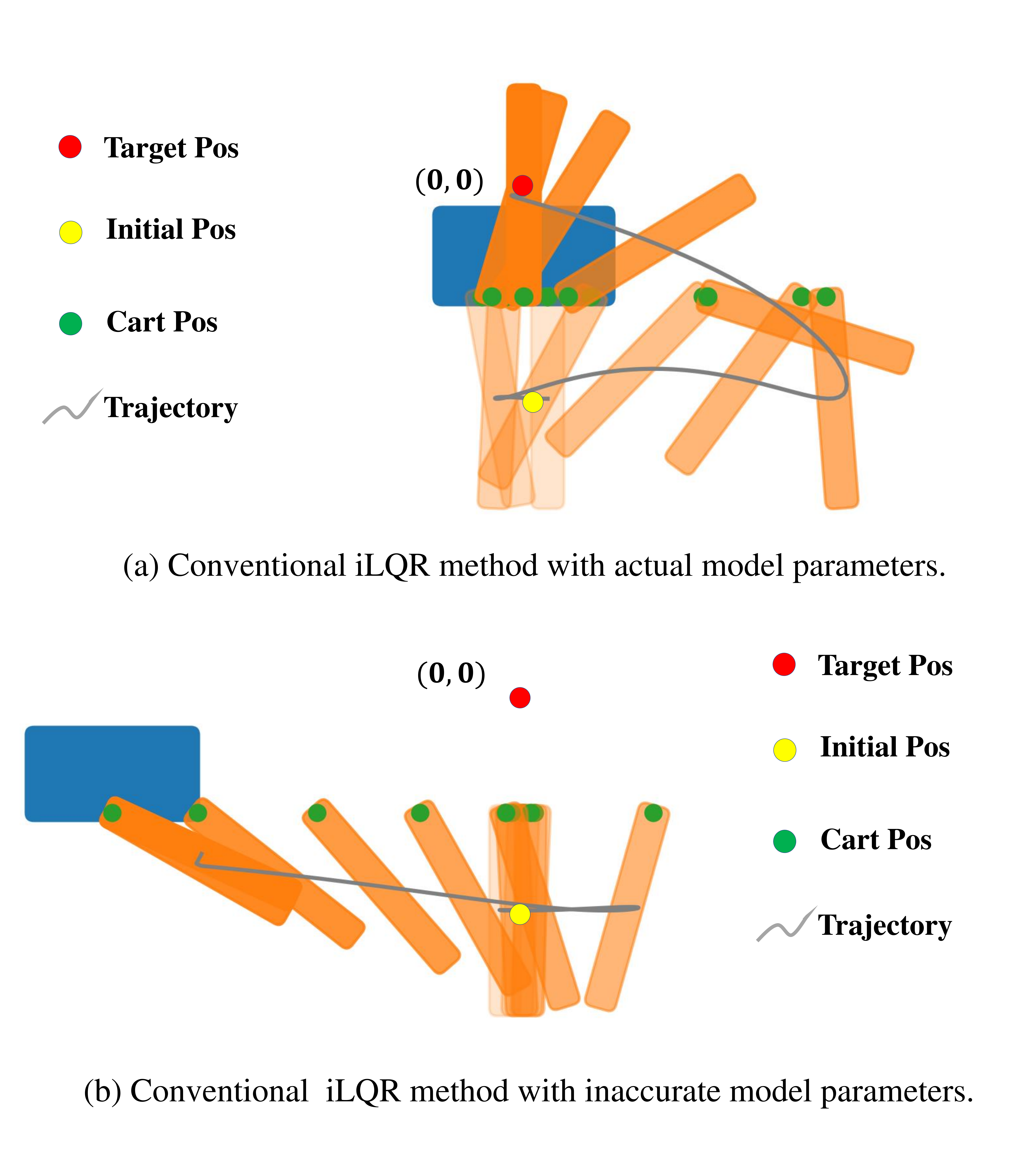}
	\setlength{\abovecaptionskip}{-10pt} 
	\caption
	{Visualization of the trajectory(middle point of the pole) generated by iLQR method in cartpole control problem using numerical simulation. In this problem, the cart moves along the horizontal line to keep the pole vertical. (a) shows the result of conventional iLQR with accurate model parameters, while (b) fails due to model inaccuracy. Opacity of the pole increases over time. For the cart, only the final position is shown while middle ones are represented by the green points for clearance.}
	\label{fig:cartpole_success}
\end{figure}

In terms of the model-based method in trajectory optimization, a second-order shooting method called differential dynamic programming (DDP) \cite{DDP} has gained popularity in the robotics community due to its high efficiency in dealing with nonlinear dynamic systems. 
In each stage, with the quadratic approximation of the system dynamics around the nominal state-input trajectory, it drives the current input trajectory towards the optimal direction. To further reduce the computation time, iterative linear quadratic regulator (iLQR) was proposed as a variant of DDP following the similar structure but using first-order approximation of the system dynamics instead. Remarkably, iLQR has been successfully applied to robot systems with precise models \cite{ilqr1,ilqr2,ma2022alternating}. However, considering complex dynamics of a robot system, the inevitable existence of model inaccuracy could lead to the deviation of obtained solution from the optimal control policy. Fig. \ref{fig:cartpole_success} shows a  typical failure case in model-based iLQR when modeling inaccuracy is introduced.

On the other hand, model-free methods have recently shown promising results in learning the system model and the optimal policy. Generally, a neural network is constructed and trained to generate the dynamic model or optimal policy in different ways. In \cite{zhao2020robust}, the control problem is modeled as a Markov Decision Process (MDP) \cite{bishop2006pattern} and optimal policy is then learned using Reinforcement Learning (RL) \cite{DPRL}. 
In order to learn long-term transition dynamics instead of step transitions, latent variables recurrent network is utilized to improve the performance in long prediction horizon \cite{ke2019learning}. Although learning-based methods outperform model-based methods in its versatility when dealing with complex control problems, deep neural networks and substantial scale of exploration samples are normally required to reach the satisfying solution.

To deal with the aforementioned drawbacks, a learning-aided shooting method named Neural-iLQR is presented for trajectory optimization over the unconstrained control space, which avoids the requirement of any prior knowledge of the system model in trajectory optimization tasks. In this approach, random trials are performed such that a dataset is collected, and then a filtered neural network is used to fit the measurement data locally in the current iteration. Subsequently, the well-trained neural network is implemented in the execution of the backward pass of iLQR method such that the trajectory optimization problem is solved iteratively. Moreover, we evaluate several critical factors pertinent to the optimization results by comparing the performance of different network structures and other related parameters. Finally, we demonstrate that the online-retraining mechanism could prevent the optimization process from trapping into the local minimum, whereby the optimality of the obtained trajectory can be further improved compared to conventional iLQR method.  

\section{RELATED WORK}
 
Similar framework utilizing iLQR has been applied on complex robot systems, and these works prominently show the high applicability of the iLQR scheme. In~\cite{mitrovic2010adaptive}, a locally weighted projection regression (LWPR) method is implemented to complete the non-parametric model identification, and then the iLQR method is performed as a feedforward controller to realize the objective of tracking. In~\cite{whole-body-mpc}, a constrained, time-varying LQR problem is solved based on the quadratic cost function and the linearized system using  Sequential Linear Quadratic (SLQ) algorithm (a continuous-time version of iLQR) \cite{SLQ-MPC} for a mobile manipulator. In general, a nominal system model is required in the implemention of these model-based methods.

 To  improve  the  robustness  toward  system modeling inaccuracy, 
 several closely related works are developed very recently. In~\cite{bechtle2020curious}, an iLQR framework called the curious iLQR is proposed, in which the system dynamics is reflected by the Bayesian modeling. In~\cite{nagariya2020iterative}, a multi-layer neural network is used to model the dynamics of off-road and on-road vehicles which is then used in iLQR controller. These methods are not sample efficient due to the large size of the built neural network. A recent work combines iLQR and RL in an augmenting way \cite{reinforced-ilqr}, which uses RL-based terminal cost and shortens the iLQR horizon adaptively in each stage to relax the requirement of an accurate system model. However, it stills requires a nominal system model to drive the optimization process.

\section{NEURAL-ILQR}

\begin{figure*}[!t]	
	\centering
	\includegraphics[trim=60 0 -20 0, width=1\linewidth]{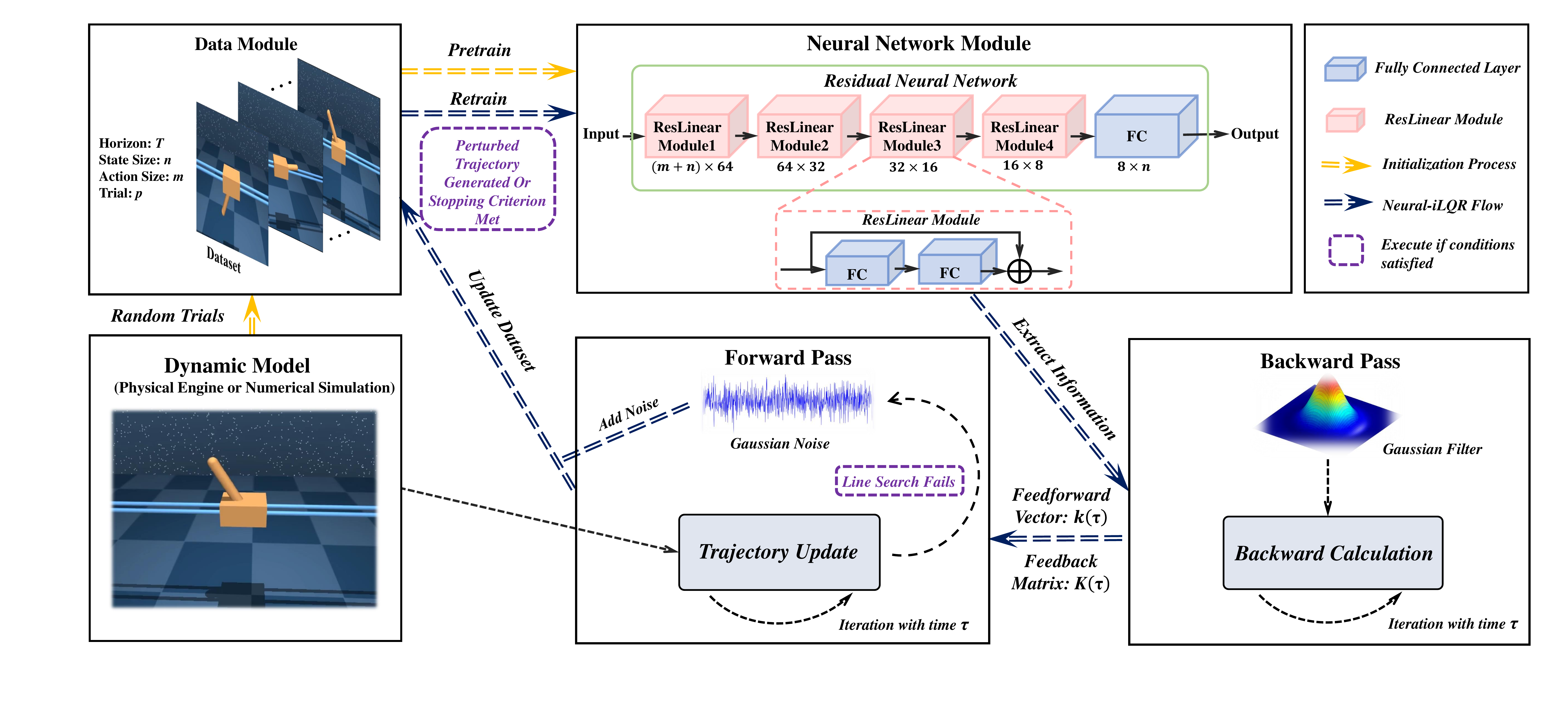}
	\setlength{\abovecaptionskip}{-14pt}
	\caption
	{Overview of the Neural-iLQR pipeline. After the initialization (yellow arrow) of the dataset and neural network, the information flow of Neural-iLQR (deep blue arrow) forms a closed loop. We use MuJoCo~\cite{todorov2012mujoco} physics engine and numerical simulation to simulate the dynamics of the system. Gaussian noise and retraining of the neural network are only needed when the criterion (purple box) are satisfied in the Neural-iLQR loop.
	}
	\label{fig:main}
\end{figure*}

\subsection{Problem Definition}
Generally, the trajectory planning problem can be expressed as a nonlinear optimization problem given in \eqref{eq:opt1}, where $x(\tau)\in\mathbb R^n$ and $u(\tau)\in\mathbb R^m$ denote the state and action at the time stamp $\tau$, respectively; 
$J_\tau\big(x(\tau),u(\tau)\big)$ denotes the cost at each time stamp $\tau$ with respective to the pair of state and action; $\phi_T$ represents the terminal cost at the time stamp $T$; $f\big(x(\tau),u(\tau)\big)$ is the dynamic function of the system which is not restricted to be linear; $x_0$ denotes the initial state of the system.
\begin{IEEEeqnarray*}{cl}~\label{eq:opt1}
	\displaystyle\operatorname*{minimize}_{\big(x(\tau),u(\tau)\big)\in\mathbb R^{n}\times \mathbb R^{m}}\quad
	&   \phi_T(x(T))+\sum_{\tau=0}^{T-1}J_\tau\big(x(\tau),u(\tau)\big) \\
	\operatorname*{subject\ to}\quad
	&x(\tau+1)=f\big(x(\tau),u(\tau)\big)\\
	&\tau=0,1,\dotsm, T-1\\
	&x(0)=x_0,\yesnumber
\end{IEEEeqnarray*}

\subsection{Overview of Neural-iLQR }


With the development of the conventional model-based iLQR, Neural-iLQR is proposed in this section with a detailed analysis to solve the optimization problem \eqref{eq:opt1}. Essentially, Neural-iLQR utilizes the neural network to learn the dynamic function and procedures of iLQR can be performed entirely based on the measurement data with the incorporation of the neural network.

As shown in Fig. \ref{fig:main}, the implementation of Neural-iLQR is summarized. In the beginning, an empty dataset is required to be initialized with $p$ random trials of prediction horizon $T$ performed in the dynamic system (i.e., the system runs arbitrarily with a sequence of control actions chosen randomly). 
Upon completion of the dataset initialization, a neural network structure can be chosen to fit the system dynamic function. 
After the setup, we conduct the backward pass and forward pass iteratively following the conventional DDP/iLQR method except for that the gradient and hessian of the dynamic function are replaced by the analytical derivatives of the neural network. Moreover, an online data collection and neural network retraining mechanisms are incorporated. 

\subsection{Backward Pass of Neural-iLQR }
The primary idea behind the backward pass of DDP/iLQR can be straightforwardly represented by the Bellman equation given by
\begin{IEEEeqnarray}{rCl}~\label{eq:Bellman}
	V_\tau\big(x(\tau)\big) = \displaystyle\operatorname*{min}_{u(\tau)} \Big\{J_\tau\big(x(\tau),u(\tau)\big)+V_{\tau+1}\Big(f\big(x(\tau),u(\tau)\big)\Big)\Big\},\nonumber\\
\end{IEEEeqnarray}
where $V_\tau(x(\tau))$ and $V_{\tau+1}(f(x(\tau),u(\tau)))$ denote the value functions with respect to the current state $x(\tau)$ and the state at the next time stamp $x(\tau+1)$, respectively.

The perturbed Q-function is then introduced \cite{ilqr2} and approximated by the second-order Taylor expansion as in \eqref{eq:pertubed-Q}, where $\delta x(\tau)$ and $\delta u(\tau)$ represent the amount of change with respect to the nominal state and action at the time $\tau$.
\begin{IEEEeqnarray}{rCl}  \label{eq:pertubed-Q}
	&&Q_\tau\big(\delta x(\tau),\delta u(\tau)\big)\nonumber\\
	&&\approx 
	\frac{1}{2}\left[\begin{array}{c}
		1 \\
		\delta x(\tau) \\
		\delta u(\tau)
	\end{array}\right]^{T}\left[\begin{array}{ccc}
		0 & \left(Q_\tau^{T}\right)_{x} & \left(Q_\tau^{T}\right)_{u} \\
		\left(Q_\tau\right)_{x} & \left(Q_\tau\right)_{x x} & \left(Q_\tau\right)_{xu} \\
		\left(Q_\tau\right)_{u} & \left(Q_\tau\right)_{ux} & \left(Q_\tau\right)_{uu}
	\end{array}\right]\left[\begin{array}{c}
		1 \\
		\delta x(\tau) \\
		\delta u(\tau)
	\end{array}\right],\nonumber\\
\end{IEEEeqnarray}
where 
\begin{IEEEeqnarray*}{rCl} \label{eq:Q_vectors_matrices_NN}
	\left(Q_\tau\right)_{x} &=&\left(J_{\tau}\right)_x+ N_{x}^\top (V_{\tau+1})_x \\
	\left(Q_\tau\right)_{u} &=&\left(J_{\tau}\right)_u+ N_{u}^\top(V_{\tau+1})_x \\
	\left(Q_\tau\right)_{xx} &=&\left(J_{\tau}\right)_{xx}+N_{x}^\top (V_{\tau+1})_{xx} N_{x}+(V_{\tau+1})_{x} \cdot N_{xx} \\
	\left(Q_\tau\right)_{ux} &=&\left(J_{\tau}\right)_{ux}+N_{u}^\top (V_{\tau+1})_{xx} N_{x}+(V_{\tau+1})_{x} \cdot N_{ux} \\
	\left(Q_\tau\right)_{uu} &=&\left(J_{\tau}\right)_{uu}+N_{u}^\top (V_{\tau+1})_{xx} N_{u}+(V_{\tau+1})_{x} \cdot N_{uu}.\\
        \yesnumber
\end{IEEEeqnarray*}

We denote the neural network that fits the system dynamic function $f$ as $N$. The optimal solution to the perturbed control action $\delta u(\tau)^*$ at the time stamp $\tau$ can then be calculated by
\begin{IEEEeqnarray}{rCl} 
	\delta u(\tau)^{*} &=&\operatorname*{argmin}_{\delta u(\tau)} \quad  Q_\tau\big(\delta x(\tau),\delta u(\tau)\big),
\end{IEEEeqnarray}
which gives
\begin{IEEEeqnarray}{rCl}\label{eq:perturbed_control_action}
	\delta u(\tau)^{*} &=& k(\tau)+K(\tau)\delta x(\tau),
\end{IEEEeqnarray}
where $k(\tau)\in\mathbb R^{m}$ and $K(\tau)\in\mathbb R^{m\times n}$ are the feedforward vector and feedback  matrix for the perturbed Q-function at the time stamp $\tau$, respectively, and they can be explicitly represented by
\begin{IEEEeqnarray*}{rCl}
	k(\tau)&=&-\left(Q_\tau\right)_{uu}^{-1} \left(Q_\tau\right)_{u} \IEEEyesnumber\IEEEyessubnumber \label{eq:policy1} \\
	K(\tau)&=&-\left(Q_\tau\right)_{uu}^{-1} \left(Q_\tau\right)_{ux}.\IEEEyessubnumber \label{eq:policy2}
\end{IEEEeqnarray*}

It is worth noting that using a neural network to learn a dynamic system with high dimensions is challenging. The gradient of the neural network model contains a large amount of noise and ineffective information compared to the gradient of the dynamic model, for which we apply a Gaussian filter to smooth the trained neural network and extract useful information. Besides, we ignore $N_{xx}, N_{ux}$ and $N_{uu}$ due to the significant noise resulted from second-order derivatives of the neural network.

\subsection{Forward Pass of Neural-iLQR }\label{section:feeforwardpass}
The last and the principal step in Neural-iLQR is the forward pass. An indispensable line search strategy is performed at the beginning of the forward pass to find a trajectory with better performance based on the given feedback matrices and feedforward vectors at different time stamps. The basic idea of the line search \cite{DDP} can be realized by
\begin{IEEEeqnarray}{rCl}
	\delta u(\tau) &=& \alpha k(\tau)+K(\tau)\delta x(\tau),
\end{IEEEeqnarray}
where $\alpha$ is the step size parameter to be determined in the line search iteration. Then the forward roll-out is conducted by feeding the perturbed action sequence to the system dynamic model $f$:
\begin{IEEEeqnarray}{rCl} \label{eq:iLQR_forward_pass}
	& u(\tau) &=\hat{u}(\tau)+\delta u(\tau)\nonumber \\
	& x(\tau+1) &=f\big(x(\tau), u(\tau)\big).
\end{IEEEeqnarray}

The nominal trajectory $(\hat x,\hat u)$ can be updated to a new feasible trajectory obtained in the forward pass, and the updated feasible trajectory can be used to initiate the next Neural-iLQR iteration and retrain the neural network.

The stopping criterion for one iteration can be chosen as the difference of the objective function value between the current and the latest nominal trajectory, once the criterion is satisfied, we will retrain the neural network on the updated dataset. For the scenario when it is still not possible to improve the dynamic system performance as the maximum number of line search iterations is reached. The solving process is considered as trapping into local minimum in this situation. We add some perturbations to the sequence of actions when generating the current trajectory, and the neural network is then retrained to escape from the local minimum. In this way, the trajectory could continue converging.

\section{Experiments}
In this section, we comprehensively evaluate Neural-iLQR using numerical simulation and MuJoCo \cite{todorov2012mujoco} physic engine with two illustrative examples: a vehicle tracking problem and a cartpole control problem. To demonstrate the practicality and effectiveness of Neural-iLQR, we use the conventional iLQR method as the benchmark to show that the proposed model-free method is comparable to or even better than the model-based counterpart. Particularly, the robustness of our method is testified when model inaccuracy is introduced. Furthermore, the influence of several critical factors in Neural-iLQR is discussed thoroughly.

\subsection{System Setups}

\subsubsection{Vehicle Tracking Problem Formulation}

The state vector of the vehicle is defined as $x=[p_x\quad p_y\quad \theta\quad v]^T$, where $p_x$ and $p_y$ denote the position of the center of the rear axis in the Cartesian coordinates; $\theta$ denotes the heading angle of the vehicle; $v$ denotes the velocity of the vehicle. The action vector is defined as $u=[\omega\quad a]^T$, where $\omega$ denotes the steering angle and $a$ denotes the acceleration. Details of the derivation of dynamic function could be refereed from \cite{DDP}.
The control target is to drive the vehicle in a straight line at a specified speed, and the reference trajectory is chosen as $r = [0\quad {-10}\quad 0\quad 8]^\top$; the objective function is chosen as a typical linear quadratic form with constant weighting matrices $Q$ and $R$ and constant reference trajectory $r$:
\begin{IEEEeqnarray*}{rCl}
J=\sum_{\tau=0}^T \Big( \big(x(\tau)-r\big)^\top Q\big(x(\tau)-r\big) + u(\tau)^\top Ru(\tau)\Big).\IEEEeqnarraynumspace\yesnumber
\end{IEEEeqnarray*}
\subsubsection{Cartpole Control Problem Formulation}
 
The state vector of the cartpole is defined as $x=[\theta \quad \omega \quad p \quad v]^\top$, where $\theta$ and $\omega$ denote the angle between the pole and the
vertical direction and its corresponding angular velocity, respectively; $p$ and $v$ denote the position and the velocity of the cart, respectively. The control action $F$ means the force applied on the cart horizontally. The system dynamic function follows \cite{cart-pole-dynamic}, and the detailed derivation is skipped here for brevity. We are going to swing up the pole and keep it balanced around the upward position by applying a force to the cart; 
the objective function is chosen as a typical linear quadratic form with the terminal state penalty:
\begin{IEEEeqnarray*}{rCl}
J= x(T)^\top Q_Tx(T) + \sum_{\tau=0}^{T-1} \Big( x(\tau)^\top Qx(\tau)+ u(\tau)^\top Ru(\tau)\Big).\\\yesnumber
\end{IEEEeqnarray*}
Noted that dynamic functions of the systems are only required for calculation in conventional model-based iLQR, and it can be fully replaced by the simulation environment and the neural network in the implementation of Neural-iLQR.

\subsection{Overall Performance of Neural-iLQR} 
\begin{figure}[h]
	\centering
	\includegraphics[trim=30 0 10 0, clip,width=0.51\textwidth]{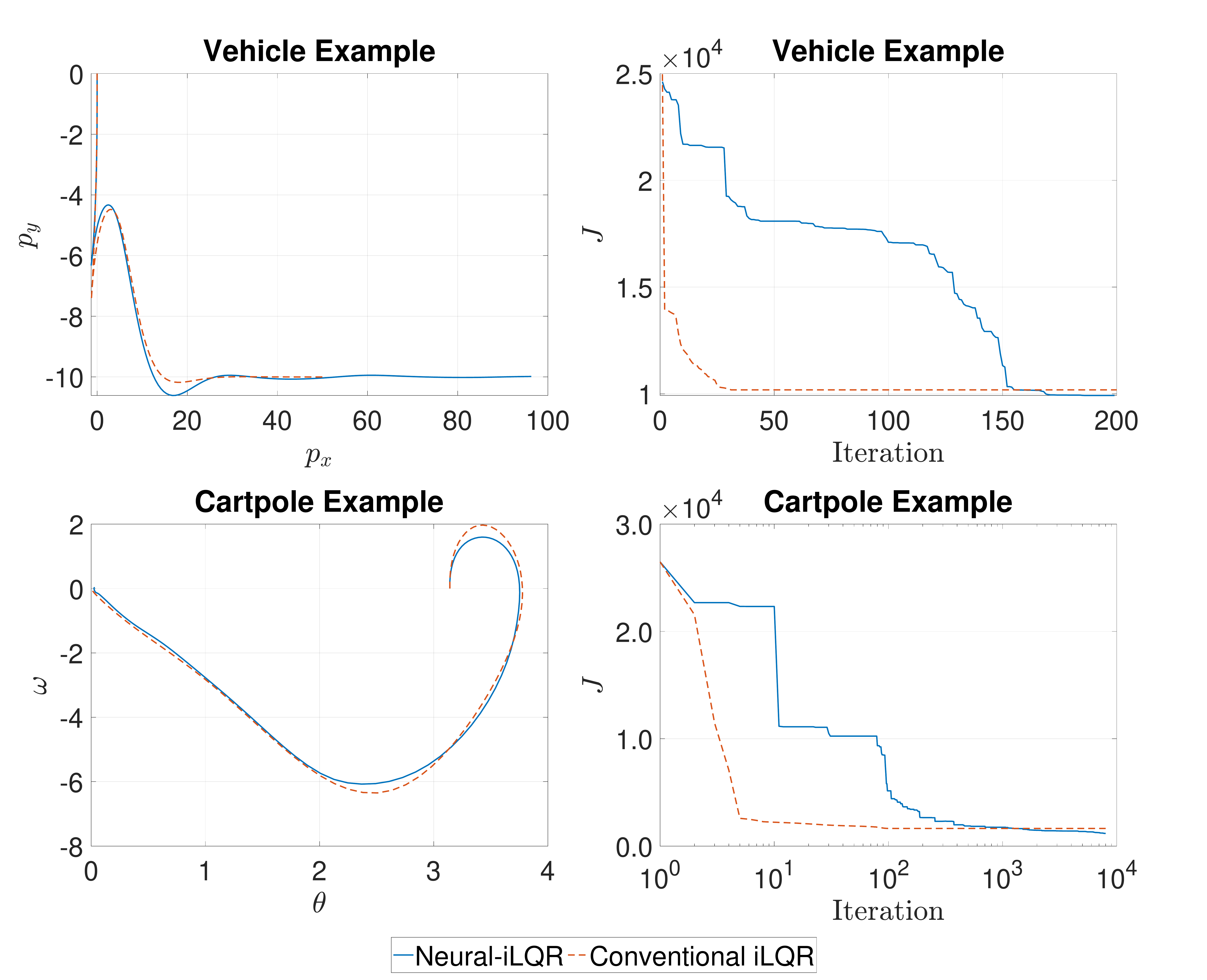}
    \setlength{\abovecaptionskip}{-14pt}
	\caption{Comparison of the trajectory and objective function value generated with Neural-iLQR and the conventional iLQR.  residual neural network is applied in this experiment, the trials number is 100, and the standard deviation of the Gaussian filter is chosen as 5.}
	\label{fig:overall}
\end{figure}

Provided with accurate dynamic model, the conventional iLQR method is proven to achieve effective optimization outcomes in such two examples indicated in Fig. \ref{fig:overall}. After several iterations, the trajectory will converge asymptotically, and the objective function is reduced to a satisfying point; thus, it is a good benchmark for comparison with our proposed model-free method. The optimal objective function values reached by the conventional model-based iLQR are ${10192}$ and ${1642}$ respectively for vehicle tracking and cartpole control examples.

As observed from Fig. \ref{fig:overall}, the proposed model-free Neural-iLQR method can successfully generate a reliable trajectory for the control problem, attaining objective function values at ${992.9}$ and ${1178}$, which shows comparable optimization performance compared to the conventional iLQR method. We apply simple neural networks to fit the local trajectory data in the current iteration and use its filtered gradient information to guide the optimization directions. When it gets stuck at some local minimum points using the current neural network, the online data
collection and retraining mechanisms discussed in Section \ref{section:feeforwardpass} empower the proposed method with the ability to avoid trapping into the local minimum. It renders possibility for the Neural-iLQR to performance even better than the conventional model-based iLQR as shown in Fig. \ref{fig:overall}.

\subsection{Comparison of Effects on Critical Factors}

The influence of the selected three critical factors in the Neural-iLQR architecture is discussed in this section, and we find that these critical factors could directly affect the optimization performance of Neural-iLQR, and thus it renders possibility to continuously improve the results by adjusting these factors in the proposed architecture.  

\subsubsection{Neural Network Architecture}

We propose three neural network structures and demonstrate the performance attained by each in this section. The first two are fully connected neural networks (FCNN) with two hidden layer and one output layer, structures are shown in Table~\ref{table:network}.
\begin{table}[h]
	\centering
	\caption{Layer Size for the Fully Connected Neural Network}
	\label{table:network}
	\begin{tabular}{|c|c|c|c|}
		\hline
		Network    & \begin{tabular}[c]{@{}c@{}}First \\ Hidden Layer\end{tabular} & \begin{tabular}[c]{@{}c@{}}Second \\ Hidden Layer\end{tabular} & \begin{tabular}[c]{@{}c@{}}Output \\ Layer\end{tabular} \\ \hline
		Small FCNN & $[(m+n)\times 128]$                                           & $[128\times 64]$                                               & $[64\times n]$                                          \\ \hline
		Large FCNN & $[(m+n)\times  1024]$                                         & $[1024\times 512]$                                             & $[512\times n]$                                         \\ \hline
	\end{tabular}
\end{table}

Shown in Fig. \ref{fig:main}, the third type of the neural network is a residual neural network. It consists of four ResLinear modules and one fully connected linear output layer. Batch normalization and ReLU are required before each fully connected layer.

We define the deviation between the objective function values obtained by Neural-iLQR and the conventional iLQR method given the dynamic model during iterations as $d$ and the iteration number when it first reaches the minimum objective function value as $k$. Neural-iLQR is randomly performed five times with each neural network structure, and the conventional iLQR method is performed with the same parameters as used in Neural-iLQR for the fair comparision.

\begin{figure}[t]
	\centering
	\includegraphics[trim=30 0 10 0, clip,width=0.48\textwidth]{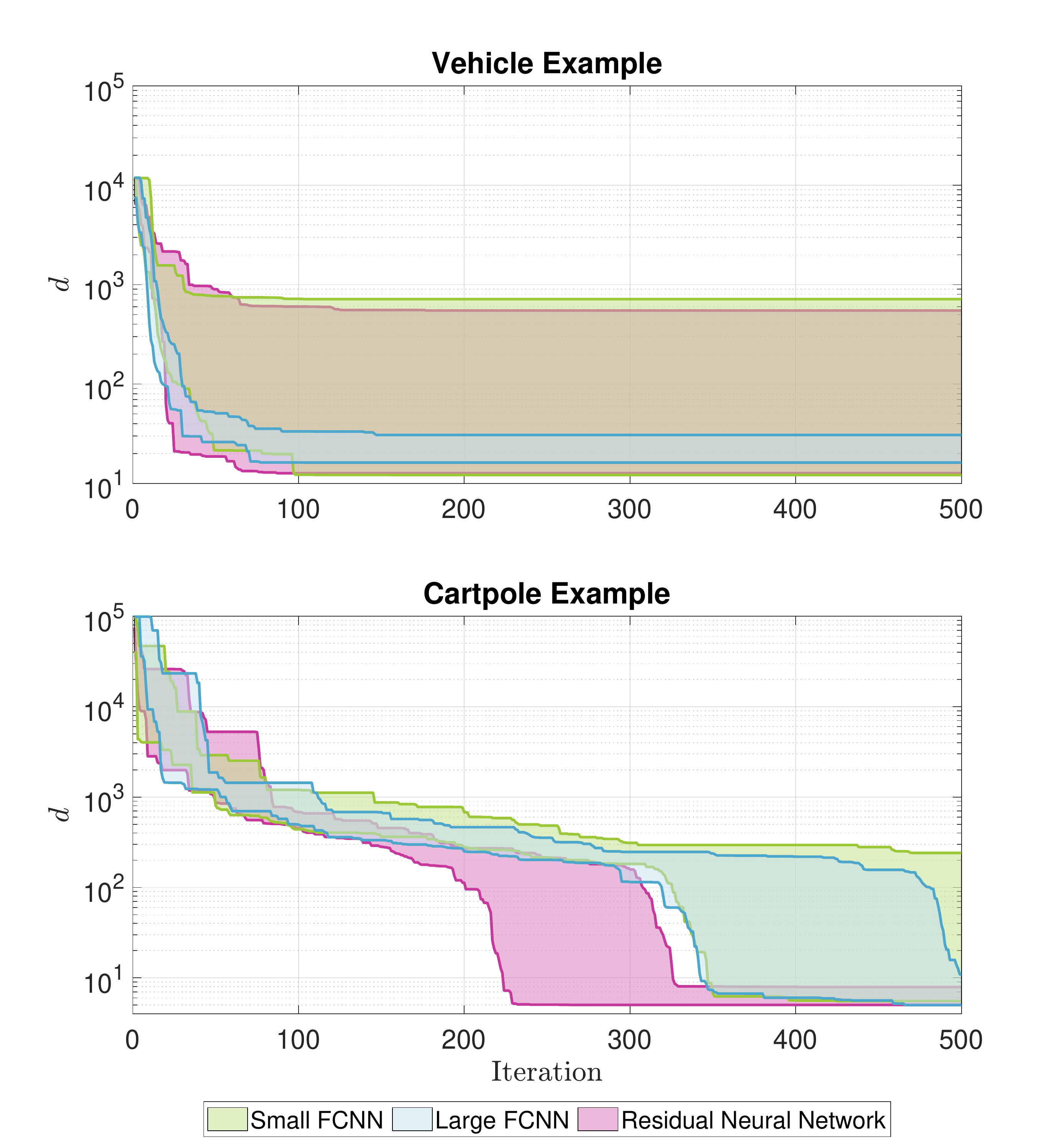}
    \setlength{\abovecaptionskip}{-14pt}
	\caption{Objective function value in iterations using Neural-iLQR with different neural network structure.}

	\label{fig:effectivenss}
\end{figure}
Fig.~\ref{fig:effectivenss} shows the deviation of the Neural-iLQR method with respect to different neural network structures after 500 Neural-iLQR iterations, respectively. From the results in Table \ref{tab:network_sturcture}, it can be seen that the large FCNN uses less time than the other two in pretraining. Moreover, Neural-iLQR method with all the three neural network structures can achieve satisfying performance in terms of the objective function value but lead to various performance. The large FCNN structure shows the highest performance in the vehicle tracking example, but the residual neural network structure demonstrates the best performance both in its optimality and fast convergence in the cartpole control example. Basically, both the large FCNN structure and  residual neural network structure provide satisfying results and show higher performance than the small FCNN structure as they can achieve better optimization results with fewer iterations. 

\begin{table}[!t]
	\scriptsize
	\centering
	\caption{Comparisons of Different Neural Network Architectures}
	\renewcommand\tabcolsep{2pt}
	\resizebox{1.0\linewidth}{!}{
	\begin{tabular}{ccccccc}
    \toprule
	  & Network Structure & Pretraining Time (s) & $d_{min}$ & $d_{avg}$ & $k_{min}$\\
	\midrule
	& Small FCNN    & 246.3256     & 13   & 295   & 98  \\
	Vehicle Tracking & Large FCNN    & \textbf{121.0006}    & 16.2    & \textbf{24}   & 77  \\
         & Residual Neural Network    & 205.6572    & \textbf{12.7}    & 125   & 87    \\
         \midrule
	& Small FCNN    & 63.7257     & 6   & 53   & 340  \\
	Cartpole Control & Large FCNN    & \textbf{28.1529}    & 5    & \textbf{7}   & 350  \\
         & Residual Neural Network    & 210.2348    & 5    & \textbf{6}   & \textbf{220}  \\ 
	\bottomrule
	\end{tabular}}
	\label{tab:network_sturcture}%
\end{table}%

\subsubsection{Gaussian Filter}
\begin{figure}[h]
	\centering
	\includegraphics[trim=10 0 80 40, clip, width=1\linewidth]{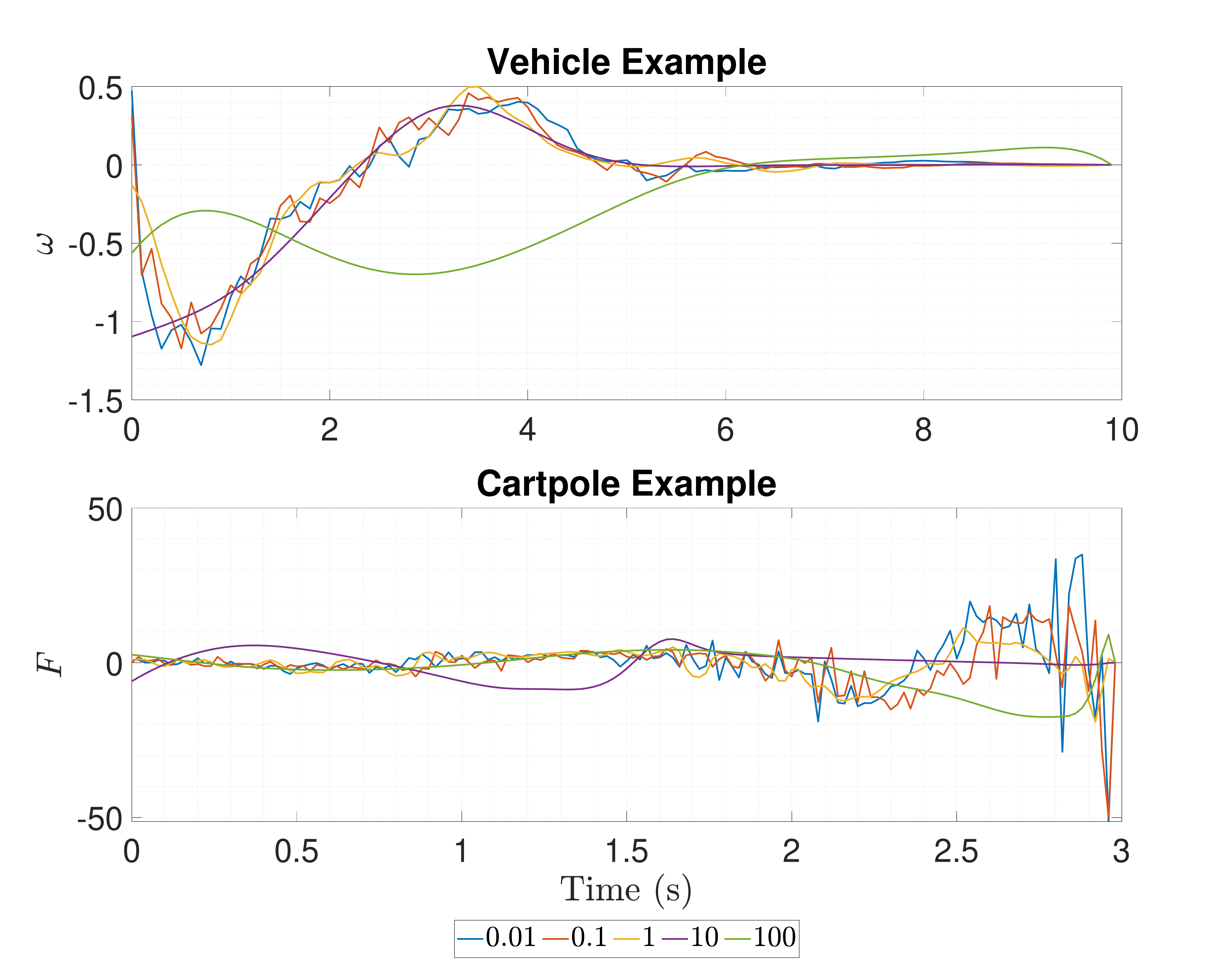}
    \setlength{\abovecaptionskip}{-14pt}	
    \caption{Control action sequence obtained by Neural-iLQR with different standard deviation parameters in the Gaussian filter.}

	\label{fig:gaussianfilterinput}
\end{figure}
Fig.~\ref{fig:gaussianfilterinput} shows the control action sequences obtained by applying Gassuian filter with different standard deviation. We can see that the smoothness of the control signals in the two illustrative examples is significantly improved with larger standard deviation of the Gaussian filter. However, to the best knowledge of the authors, an extremely large standard deviation parameter also leads to unsatisfying optimization results due to the lack of sudden changes in control actions.

\subsubsection{Number of Trials in Dataset}
\begin{figure}[h]
	\centering
	\includegraphics[trim=10 0 80 40, clip, width=1\linewidth]{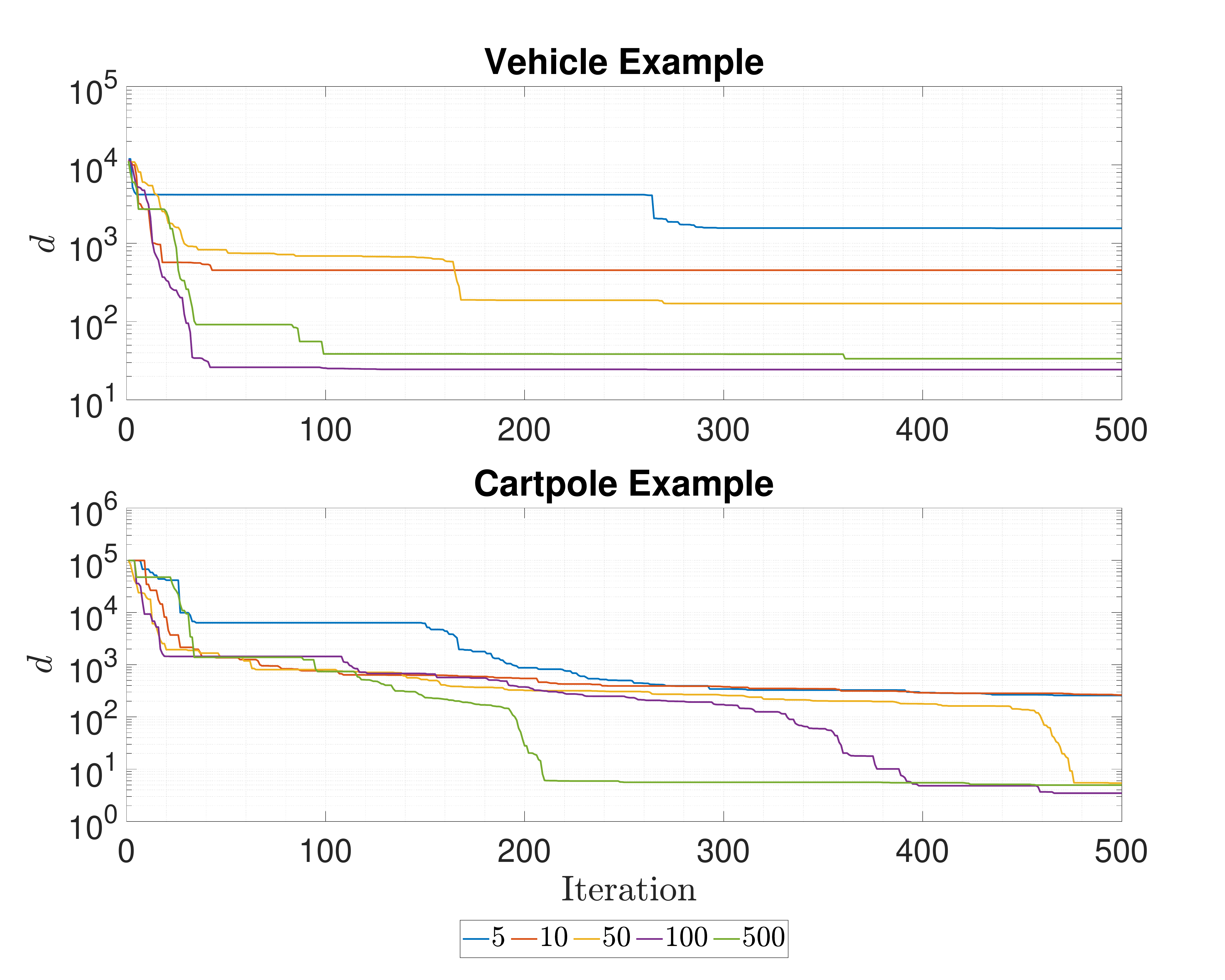}
    \setlength{\abovecaptionskip}{-14pt}	
    \caption{Deviation of the objective function value from the conventional iLQR method by using Neural-iLQR with different number of trials.}
	\label{fig:trials}
\end{figure}

\begin{figure}[h]
	\centering
	\includegraphics[trim=20 0 20 0, width=0.95\linewidth]{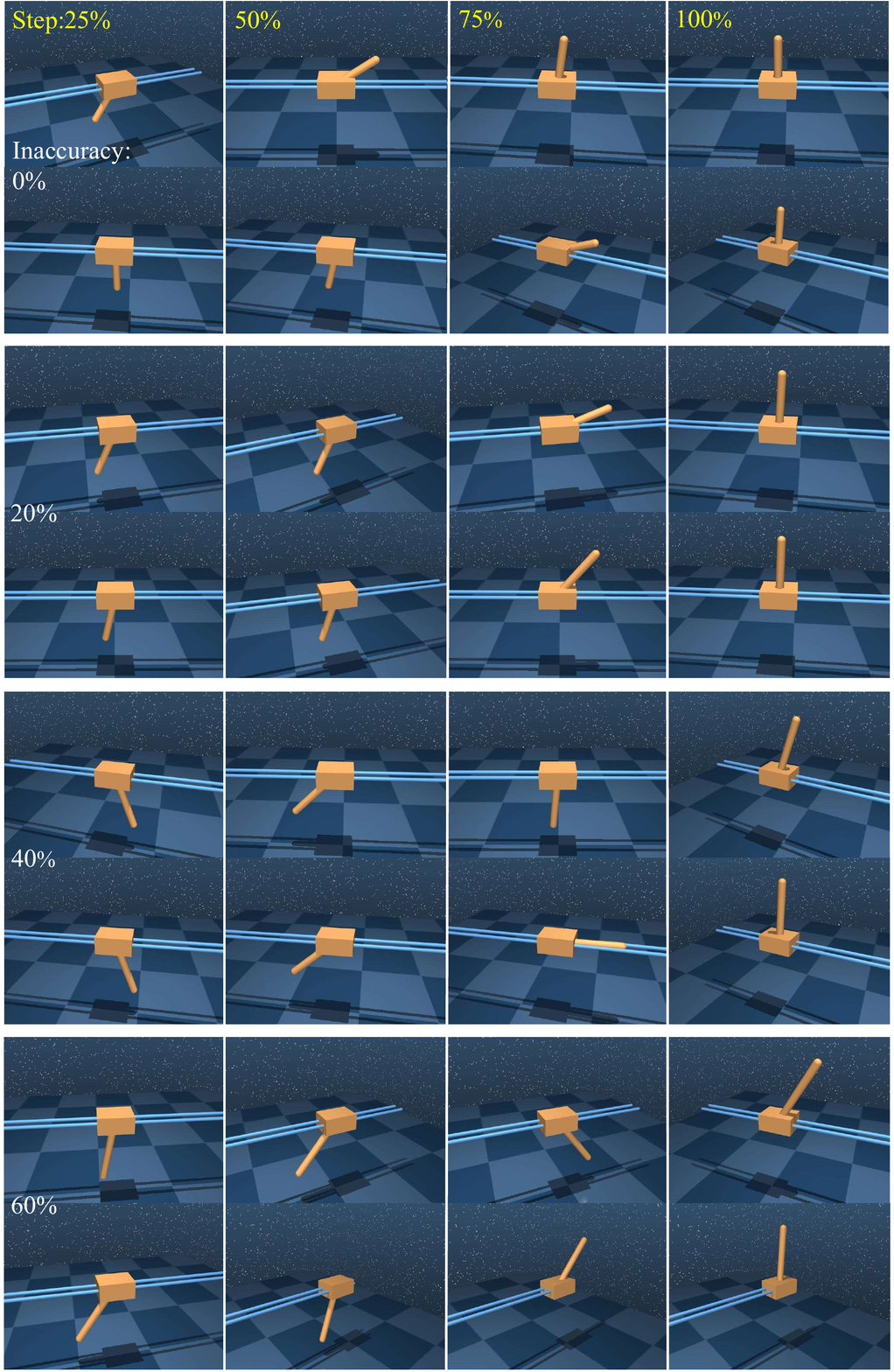}
 \setlength{\abovecaptionskip}{-2pt}	
    \caption{Visualization of the comparisons between the conventional model-based iLQR (above) and the proposed Neural-iLQR (below) under different level of model inaccuracy.}
	\label{fig:comparison_mujoco}
\end{figure}

Fig.~\ref{fig:trials} shows the deviation of Neural-iLQR with different numbers of trials. From Fig.~\ref{fig:trials}, it shows that the convergence of the obtained trajectory could usually be improved with larger number of trials. However, an extremely large number of trials can lead to difficulty in training the neural network. The experiments in the two examples show that it is usually rational to choose the number of trials to be around 100.

\begin{table}[h]
	\centering
        \renewcommand\arraystretch{1.2}
	\caption{Comparison of Neural-iLQR and model-based iLQR for cartpole control experiment}
	\label{table:MuJoCo}
        \resizebox{1\linewidth}{!}{
	\begin{tabular}{|c|c|c|c|c|c|c|}
		\hline
		\multirow{2}{*}{Model Inaccuracy} & \multicolumn{3}{c|}{Model-based iLQR}  &  \multicolumn{3}{c|}{Neural-iLQR}  \\ \cline{2-7} 
		& Success &$\theta_{error}$ & Obj.Val ($\times 10^3$) & Success & $\theta_{error}$ & Obj.Val ($\times 10^3$)\\ \cline{1-7}
 		0\% & Yes  & \textbf{4.750}   & \textbf{2.334} & Yes    & 7.051   & 3.269\\ 
		20\%   & Yes    & 7.872    & 3.436 & Yes      & \textbf{6.965}     & \textbf{2.477} \\ 
 		40\% &  No  & 9.456   & 9.833 & Yes    & \textbf{7.984}   & \textbf{3.458}\\
		60\%   &  No    & 9.287    & 12.699 & Yes      & \textbf{7.657}     & \textbf{3.038}  \\ \hline    
	\end{tabular}}
\end{table}

\subsection{Robustness to Modeling Inaccuracy}

In this section, we further demonstrate the robustness of the proposed method against model inaccuracy compared to the conventional iLQR method in MuJoCo. We build a cartpole model and use the simulation model to conduct forward dynamics by feeding inputs to the environment. We can obtain the trajectory and collect the real-time data for training in Neural-iLQR, which is convincing and close to the scenarios in the real world. Modeling inaccuracy is inevitable in the real-world situation, in this case, we introduce the model inaccuracy to the system by adjusting the model parameter in MuJoCo.

As indicated in Fig. \ref{fig:comparison_mujoco} and Table \ref{table:MuJoCo}, the conventional model-based iLQR shows its effectiveness in achieving satisfying optimization results with the accurate model. The objective function value reaches ${2.334}$ and the mean square error (MSE) of $\theta$ for the generated trajectory is ${4.750}$. However, its optimization performance will be significantly affected by the modeling inaccuracy and it may even fail when large inaccuracy is introduced. Meanwhile, we can see from the results in Table \ref{table:MuJoCo} that the proposed Neural-iLQR method shows comparable capability in generating the optimal trajectory towards the control target without any prior knowledge of the system model, and its robustness and adaptability to inaccurate model is demonstrated.

\section{CONCLUSION}
This paper investigates the development of Neural-iLQR, a learning-aided shooting method for trajectory optimization. In view of an unknown dynamic system, a neural network is utilized to fit the dynamic function in an iterative framework, which enables the use of the iLQR method in trajectory planning problems. The estimated gradient matrix of the dynamic function is derived, and the improved feedforward iteration is proposed to deal with the inaccuracy and imprecision in the optimization problem. As a result, the refined iLQR method can be applied completely without any prior information of the dynamic system. Moreover, the trajectory resulted from the Neural-iLQR method can be even better than the conventional iLQR method, as the local optimal point can be escaped with the deployment of the further exploration procedure. Finally, illustrative examples are used to validate the performance of the proposed Neural-iLQR method and detailed discussions are presented. It is worthwhile to highlight that due to the effectiveness and the universality of the proposed architecture, the framework could be suitably adjusted or extended to address practical issues in many real-world applications.
\bibliographystyle{IEEEtran}
\normalem
\bibliography{ref}

\end{document}